\documentclass{article}

\usepackage{tabularx, booktabs}
\usepackage{times}
\usepackage{graphicx}
\usepackage{sidecap}
\usepackage[small]{caption}
\usepackage{subcaption}
\usepackage{amsmath}
\allowdisplaybreaks
\usepackage{amsthm}

\usepackage{thmtools}
\usepackage{thm-restate}

\usepackage[round]{natbib}

\usepackage{appendix}

\newcommand{\eqcomment}[1]{\addtocounter{equation}{1}\tag*{\rightcomment{#1}\quad(\theequation)}}
\usepackage{suffix}
\WithSuffix\newcommand\eqcomment*[1]{\tag*{\rightcomment{#1}}}

\usepackage[hidelinks]{hyperref}

\usepackage[preprint]{neurips_2023}

\usepackage{multirow}
\usepackage{caption}
\usepackage{graphicx}
\usepackage{amsmath}
\usepackage{wrapfig}
\usepackage{amsmath,amsfonts,bm}

\usepackage[utf8]{inputenc} 
\usepackage[T1]{fontenc}    
\usepackage{hyperref}       
\usepackage{url}            
\usepackage{booktabs}       
\usepackage{amsfonts}       
\usepackage{nicefrac}       
\usepackage{microtype}      
\usepackage{xcolor}         

\usepackage{times}
\usepackage{bm}
\usepackage{soul}
\usepackage{amsthm}
\usepackage{algorithm}
\usepackage{enumitem}

\usepackage{multirow}
\usepackage[noabbrev,capitalize]{cleveref}
\crefname{equation}{equation}{equations}
\crefname{section}{section}{sections}
\crefname{footnote}{footnote}{footnotes}   
\crefname{line}{line}{lines}

\creflabelformat{section}{\color{red}#2#1#3}
\creflabelformat{figure}{\color{red}#2#1#3}
\creflabelformat{table}{\color{red}#2#1#3}
\creflabelformat{appendix}{\color{red}#2#1#3}
\creflabelformat{equation}{\color{red}#2#1#3}

\usepackage{hyperref}

\usepackage{algorithm}
\usepackage[noend]{algpseudocode}

\renewcommand\algorithmicthen{:}
\newcommand{\rightcomment}[1]{\(\triangleright\) {\small \it #1}}
\algnewcommand{\IfThen}[2]{\State \algorithmicif\ #1\ \algorithmicthen\ #2}
\algnewcommand{\IfThenElse}[3]{\State \algorithmicif\ #1\ \algorithmicthen\ #2\ \algorithmicelse\ #3}
\algrenewcommand{\algorithmiccomment}[1]{\hfill \rightcomment{#1}}
\algnewcommand{\LineComment}[1]{\State \rightcomment{#1}}
\algnewcommand{\LinesComment}[1]{\State \rightcomment{\parbox[t]{\linewidth-\leftmargin-\widthof{\(\triangleright\) }}{#1}}\smallskip}
\algnewcommand\algorithmicinput{{\bfseries Input:}}
\algnewcommand\INPUT{\item[\algorithmicinput]}
\algnewcommand\algorithmicoutput{{\bfseries Output:}}
\algnewcommand\OUTPUT{\item[\algorithmicoutput]}

\newcommand{\ourmodel}{\textsc{MixPro}\xspace}
\newcommand{\tif}[1]{\textit{\textbf{#1}}}
\usepackage{xspace}
\usepackage[algo2e]{algorithm2e}
\usepackage{adjustbox}

\usepackage{mathtools}
\usepackage{wrapfig}

\usepackage{graphicx}

\usepackage{tabularx, booktabs}
\newcolumntype{C}{>{\centering\arraybackslash}X}
\newcolumntype{R}{>{\raggedleft\arraybackslash}X}
\newcolumntype{S}{>{\raggedleft\arraybackslash\hsize=.5\hsize}X}

\usepackage{tikz}
\newcommand*{\circled}[1]{\lower.7ex\hbox{\tikz\draw (0pt, 0pt)%
    circle (.5em) node {\makebox[0.1em][c]{\small #1}};}}

\title{\textsc{MixPro}\xspace: Simple yet Effective Data Augmentation for Prompt-based Learning}

\author{
  Bohan Li$^1$, Longxu Dou$^1$, Yutai Hou$^1$, Yunlong Feng$^1$, Honglin Mu$^1$ \\
  \textbf{Qingfu Zhu$^1$, Qinghua Sun$^{2,3}$, Wanxiang Che$^{1\spadesuit}$}\\
$^1$ Research Center for Social Computing and Information Retrieval \\ Harbin Institute of Technology, China \\
$^2$ Jilin Kexun Information Technology Co., Ltd., Beijing, China\\
$^3$ iFLYTEK Research, Beijing, China\\
\texttt{\{bhli, lxdou, ythou, ylfeng, hlmu, qfzhu\}@ir.hit.edu.cn} \\
\texttt{qhsun2@iflytek.com}, \texttt{car@ir.hit.edu.cn}
}

\begin{document}

\maketitle
\def\thefootnote{$\spadesuit$}\footnotetext{Corresponding author.}
\def\thefootnote{\arabic{footnote}}
\begin{abstract}

  Prompt-based learning has shown considerable promise in reformulating various downstream tasks as cloze problems by combining original input with a predetermined template.
  This approach demonstrates its effectiveness, especially in few-shot learning scenarios, where the model is trained on a scarce amount of data.
  Despite its successes, the limited templates and text in few-shot prompt-based learning scenarios leave significant room for performance improvement.
  Moreover, existing methods sometimes resort to model ensembles, which, while effective, could potentially hamper model efficiency due to increased computational demands~\citep{schick-schutze-2021-just}.
  To address these issues, we introduce \ourmodel, an augmentation method designed to augment both the vanilla input text and the templates.
  We implement this through the token-level, the sentence-level, and the template-level Mixup strategies.
  The experimental results on five few-shot datasets show that \ourmodel outperforms other augmentation baselines, improving model performance by an average of $5.08\%$ compared to before augmentation. 
  
\end{abstract}

\section{Introduction}

Prompt-based learning has recently received significant attention~\citep{DBLP:journals/corr/abs-2107-13586}.
This approach reformulates downstream tasks as cloze questions using a prompt template~\citep{schick-schutze-2021-just}.
Hard prompts, which use an actual text string suitable for human reading as the template, have demonstrated excellent performance in a variety of downstream tasks~\citep{DBLP:conf/nips/BrownMRSKDNSSAA20,gao-etal-2021-making,seoh-etal-2021-open,qi-etal-2022-enhancing,zhong-etal-2021-adapting-language,10.1162/tacl_a_00468}.
The templates play a crucial role in prompt-based learning, as pre-trained language models output labels by utilizing them to transform different tasks into statistical cloze questions~\citep{DBLP:conf/naacl/ClarkLCK0T19}.
Prompt-based learning can also be applied to \textit{few-shot} learning settings, where the model can leverage information from a small number of training instances directly~\citep{schick-schutze-2021-just}.

\begin{figure}[h]
  \centering 
  \includegraphics[width=1.0\linewidth] {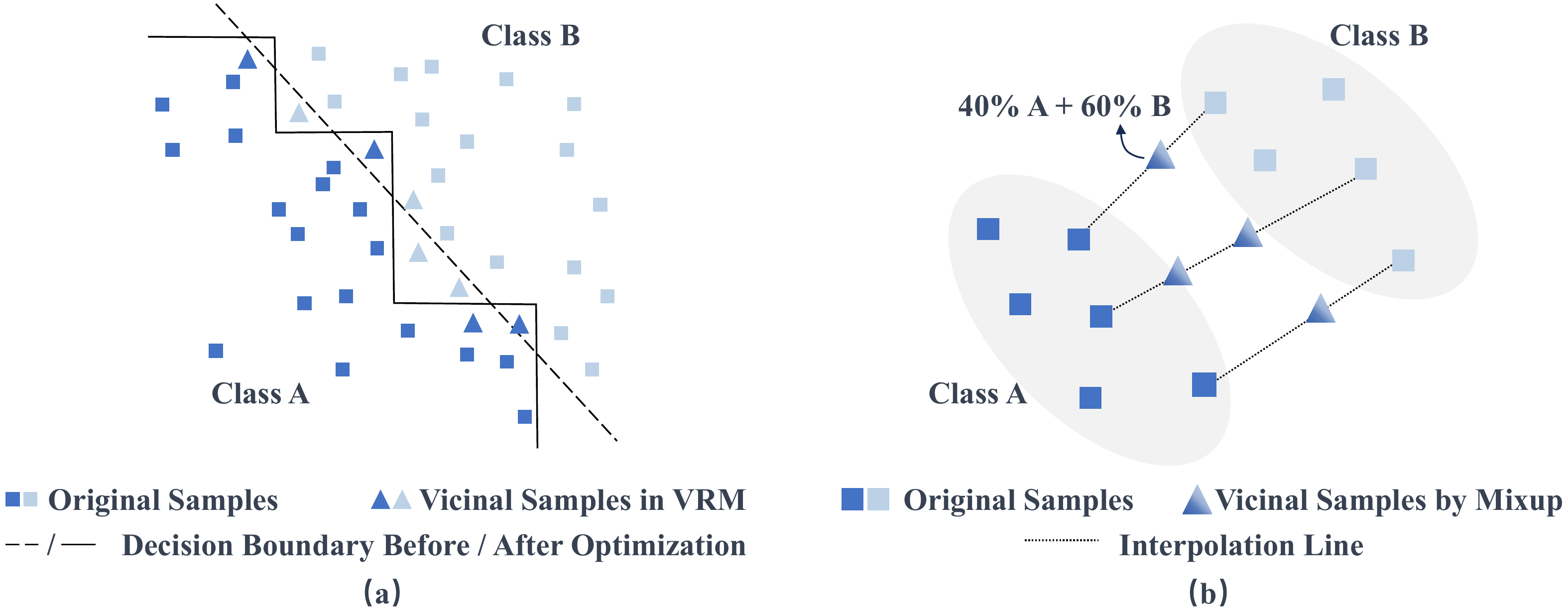}
  \caption{ 
(a) The purpose of VRM is to optimize for both the original samples and the vicinal samples, enhancing the model's robustness and generalization capability. (b) Mixup can be viewed as an implementation of VRM, creating vicinal samples by interpolation.}
  \label{fig:intro}
\end{figure}

However, the performance of few-shot prompt-based learning models can be sensitive to templates variation~\citep{gao2020making,vandeKar2022DontPS}, with small changes causing significant performance drops~\citep{cao-etal-2022-prompt}.
Moreover, existing methods~\citep{schick-schutze-2021-just} using model ensembles with different templates would constrain the model inference efficiency.
Furthermore, model performance on Natural Language Processing (NLP) tasks is often heavily dependent on the size of the training data~\citep{ren-etal-2021-text}. 
Therefore, we argue that the limited number of samples and templates in few-shot prompt-based learning significantly limits model performance.

To address this challenge, we intend to adopt a data augmentation (DA) strategy~\citep{Chen2022DecouplingKF}, enhancing both vanilla input text and templates by synthesizing additional distinct samples based on original samples~\citep{Simard2000TransformationII,Szegedy2014GoingDW,Wei2019EDA[5],li2022data}.
DA methods have demonstrated their superiority in many scenarios, particularly in low-resource settings where annotations are difficult and require a significant amount of expert knowledge or experience~\citep{Hu2019LearningDM,anaby2019not[20], Chen2021AnES}.
Vicinal Risk Minimization (VRM)~\citep{chapelle2000vicinal} offers significant inspiration in this regard. Its core principle is to optimize models not just on the original data but also in the \textit{vicinal} regions surrounding that data (Fig.~\ref{fig:intro} (a)). This approach bolsters the model's resilience to minute data perturbations.

Inspired by VRM, in this paper, we introduce the Mixup strategy~\citep{ZhangCDL18_mixup} into few-shot prompt-based learning.
As an implementation strategy of VRM, Mixup linearly blends two training samples and their respective labels, effectively generating new vicinal samples within the data space (Fig.~\ref{fig:intro}(b)).
Specifically, our method is a comprehensive three-level \textbf{Mix}up for \textbf{Pro}mpt-based learning (\ourmodel), including the token-level, the sentence-level, and the template-level Mixup strategies.
It blends word embeddings, mixes hidden [MASK] prompt representations, and uses diverse templates during training.

We conduct experiments on five Natural Language Understanding (NLU) datasets and demonstrate that \ourmodel consistently outperforms previous DA baselines.
It improves model performance by an average of $5.08\%$ compared to before augmentation, with specific gains of $4.20\%$, $9.76\%$, and $6.49\%$ on CB~\citep{Marneffe2019TheCI}, RTE~\citep{DBLP:conf/mlcw/DaganGM05,BarHaim2006TheSP}, and BoolQ~\citep{DBLP:conf/naacl/ClarkLCK0T19} respectively.
Notably, \ourmodel is more efficient, using only $1/n$ inference time, where $n$ denotes the template count of the corresponding task. 
We also conduct ablation experiments to analyze the contributions of three-level Mixup strategies, the necessity of augmenting both text and templates, and the fluctuation of model performance.

Our contributions are as follows:

\begin{itemize}
\item We systematically investigate the \textbf{challenges} of few-shot prompt-based learning augmentation, which include high sensitivity to templates, limited input text and templates, and inefficient inference time cost.
\item We introduce MixPro, a \textbf{comprehensive} DA framework for the entire prompt, which encompasses both the vanilla input text and the templates with three-level Mixup strategies.
\item Our proposed method achieves the \textbf{best performance} compared to other augmentation baselines across five NLU datasets on average and significantly improves model performance by \textbf{$5.08\%$}.
\end{itemize}

\section{Background}\label{Sec.Background}

We start by briefly introducing the basics of prompt-based learning.
Then, we introduce the vanilla Mixup strategy~\citep{guo2019augmenting[24]}, which serves as our baseline.

\subsection{Problem Definition: Prompt-based Learning}\label{model:Prompt-based-Learning}

Pre-trained language models (PLMs) like BERT~\citep{devlin-etal-2019-bert} and ALBERT~\citep{DBLP:conf/iclr/LanCGGSS20} have become essential in NLP, showing proficiency in various tasks. Recent studies~\citep{brown2020language,schick-schutze-2021-just} demonstrate their ability for few-shot learning, transforming input text into prompts via templates for training without parameter updates, utilizing only a few support instances~\citep{DBLP:journals/corr/abs-2107-13586}.

Let's take PET~\citep{schick-schutze-2021-just} as an example, which is a popular work in prompt-based learning and also serves as our backbone. In the sentiment analysis task, we are given a vanilla input text $x$ =``This movie is amazing.'' and we can choose a template $t$ = ``\textit{The feedback is [MASK]}.'' to construct a prompt $\mathbf{p}$ as follows:
\begin{eqnarray}
  \mathbf{p} = \mbox{\small{This movie is amazing. \textit{The feedback is [MASK]}}}.
\label{prompt-example}
\end{eqnarray}

We then ask the pre-trained language model to fill in the [MASK] symbol with a word (e.g., ``positive'' or ``negative'') from a specific set $W$. The chosen word $w \in W$ is mapped to a label through an injective function called a verbalizer $v: W \rightarrow L$, where $L$ is the label set of the corresponding task. This label serves as the final prediction of the model.

\subsection{Why Augment Few-shot Prompt-based Learning?}
Large Language Models (LLMs) such as GPT-3~\citep{brown2020language}, LLAMA-2~\citep{Touvron2023Llama2O}, and PaLM-2~\citep{Anil2023PaLM2T} enable remarkable performances across a wide range of tasks.
Additionally, they perform well even in zero/few-shot settings~\citep{Yuan2023DistillingSK}.
However, they can sometimes struggle in specialized domains with scarce data, including fields like science~\citep{Cohan2019StructuralSF}, psychology~\citep{Wang2023CognitiveDB}, and medicine~\citep{Dernoncourt2017PubMed2R}, as these models largely rely on general corpora.
Moreover, training LLMs in low-resource settings can be both challenging and costly 
due to 
their vast number of parameters complicates the training process~\citep{Yang2023Baichuan2O}.
There is a data shortage issue, and training such models requires high costs and demands advanced techniques~\citep{Yuan2023DistillingSK}.
These demands make them less practical in certain situations.

In contrast, in low-resource scenarios, smaller and specialized PLMs (e.g., BERT~\citep{Devlin2019BERTPO}, ALBERT~\citep{Lan2019ALBERTAL}) are good choices for rapid adaptation to downstream tasks~\citep{Yuan2023DistillingSK,Li2023ThinkOT}.
This can be achieved through \textbf{few-shot prompt-based learning} to transform downstream tasks into prompts given some annotated templates~\citep{schick-schutze-2021-exploiting}.
This technology provides adaptability to new tasks with minimal task-specific data, is computationally efficient, saves resources, and 
leverages vast knowledge from PLMs.

Few-shot prompt-based learning, while promising, faces three key challenges. 
\tif{(1)} Its performance is greatly affected by template and training example choices, leading to significant drops with minor changes~\citep{Nie2022ImprovingFP}. 
\tif{(2)} Using multiple templates for training and ensembling models during inference is necessary, but inefficient for practical applications.
\tif{(3)} The scarcity of templates and text in few-shot prompt-based learning restricts model accuracy.
Addressing these challenges can enhance few-shot prompt-based learning model performance.

In this paper, we argue that the limited number of samples and templates in few-shot prompt-based learning is the core reason.
Building upon this foundation, data augmentation techniques can mitigate the high cost of manual annotation, thereby automating further improvements in model performance~\citep{li2022data}.

\begin{figure}[h]
  \centering 
  \includegraphics[width=1.0\linewidth] {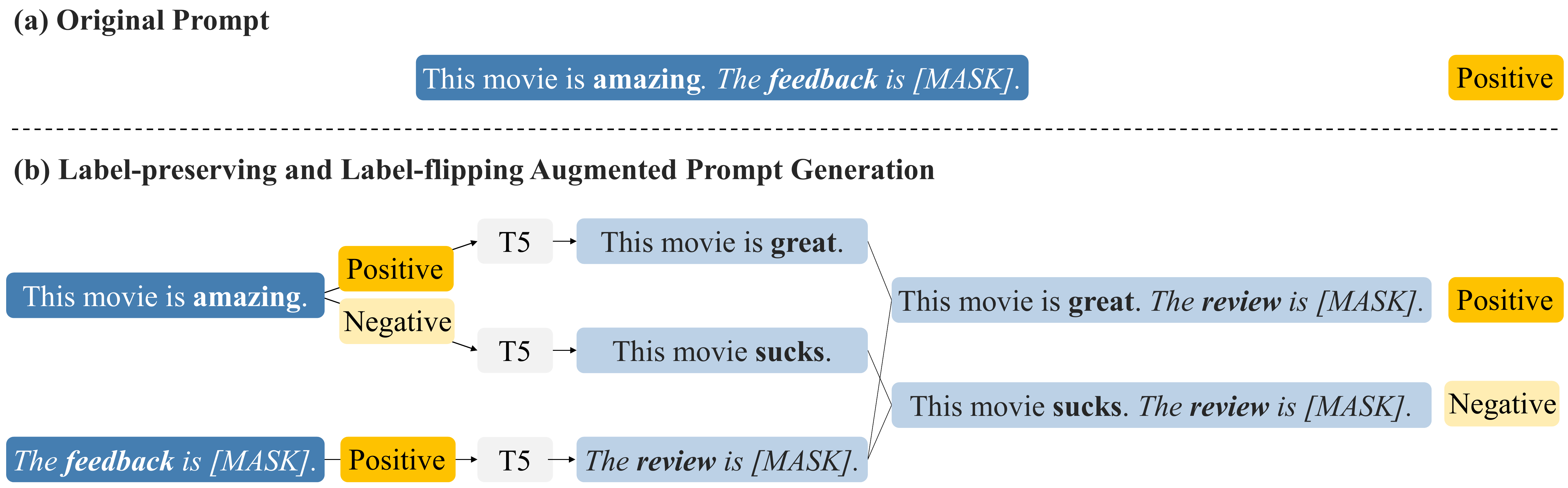}
  \caption{ 
  Based on the original prompt (a), we generate both label-preserving and label-flipping prompts through T5 (b).
  The augmentation is applied to the entire prompt, including the vanilla input text and the \textit{template} (in \textit{italics}). 
}
  \label{fig:DA_example}
\end{figure}

\subsection{The Mixup Strategy}\label{Sec:background-mixup}

The Mixup strategy expands training distribution by creating \textit{virtual} examples from nearby examples.
By encouraging the model's linear behavior between examples, Mixup minimizes erratic predictions beyond the training set~\citep{ZhangCDL18_mixup}, yielding a more robust and generalizable model.

To apply the Mixup strategy, a pair of samples $(x, y)$ and $(x', y')$, where $x$ and $x'$ denote the vanilla input text, and $y$ and $y'$ are their one-hot labels, are chosen. Mixup then minimizes the sample loss from their vicinity distribution.
The vanilla Mixup strategy is defined as follows:
\begin{eqnarray}
x_{mixup} = \lambda \cdot x + (1 - \lambda) \cdot x', \label{equ:mixup} \\
y_{mixup} = \lambda \cdot y + (1 - \lambda) \cdot y',
\label{equ:mixup-label}
\end{eqnarray}
where $\lambda$ is the Mixup ratio drawn from a Beta distribution $\lambda \sim \beta(\alpha, \alpha)$, with $\alpha$ being a hyperparameter controlling the degree of mixing. The Mixup strategy generates a new synthetic sample $(x_{mixup}, y_{mixup})$ for model training. By interpolating between pairs of training examples, Mixup encourages the model to learn more smoothly and generalize better to unseen data.

To enhance model robustness, mixing original samples with their \textit{label-preserving} or \textit{label-flipping} augmentations is common, enhancing the distribution and boosting performance~\citep{cheng2020advaug[25], DBLP:conf/iclr/HendrycksMCZGL20}. 
We follow a similar operation in our \ourmodel~\cite{zhou2021flipda}. Given the original prompt ($\mathbf{p} = x,t$) in Sec.\ref{model:Prompt-based-Learning}, we apply T5~\citep{DBLP:journals/jmlr/RaffelSRLNMZLL20} to generate label-preserving and label-flipping augmented text, and label-flipping augmented text only.
We provide an example in Fig.~\ref{fig:DA_example}.
The label-preserving augmented prompt is ``The movie is great. \textit{The review is [MASK]} '', and the label-flipping augmented prompt is ``The movie sucks. \textit{The review is [MASK]}''.
\footnote{We systematically compare the baselines with the proposed \ourmodel from five perspectives in Section~\ref{sec:baseline}.}
\footnote{Sec.~\ref{sec:experiment-parameters} introduces more details of label-preserving and label-flipping prompts generation.}

\begin{algorithm}[th]
  \footnotesize 
  \SetAlgoLined
  \textbf{Input}:
  The original dataset $D = {(x, y)}$ and augmented dataset $D' = {(x', y')}$, original template set $T$, augmented template set $T'$, hyperparameter $\alpha$, and a masked language model $M$.\\
  \textbf{Output}:
  The cross-entropy loss $\mathcal{L}$
  
  \For{each epoch in training epochs}{
      \textbf{// Template-level Mixup} \\
      Sample an original and its augmented prompt $t$ and $t'$ from $T$ and $T'$ respectively. \\
      \For{each original and augmented sample $(x, y)$ and $(x', y')$ in $D$ and $D'$}{
          $\mathbf{p} \gets (x, t); \mathbf{p}' \gets (x', t')$; \\
          $\lambda \gets \beta(\alpha, \alpha)$; \\
          \textbf{// Token-level Mixup} \\ %
          $E_{\mathbf{p}}, E_{\mathbf{p}'} \gets$ the input representations of $\mathbf{p}$ and $\mathbf{p}'$ by Equ.~\ref{equ:emb_origin_input}-\ref{equ:emb_aug_input};\\
          $E_{mixup} \gets \lambda \cdot E_{\mathbf{p}} + (1-\lambda) \cdot E_{\mathbf{p}'}$; \\
          \textbf{// Sentence-level Mixup} \\
          $H_{\mathbf{p}}, H_{\mathbf{p}'} \gets$ the hidden vectors of $M(E_{mixup})$ at the [MASK] positions in $\mathbf{p}$ and $\mathbf{p}'$; \\
          $H_{mixup} \gets \lambda \cdot H_{\mathbf{p}} + (1-\lambda) \cdot H_{\mathbf{p}'}$; \\
          $logits \gets \mbox{MLP}(H_{mixup})$; \\
          $y_{mixup} \gets \lambda \cdot y_p + (1 - \lambda) \cdot y_{p'}$; \\
          $\mathcal{L} \gets L_{CE}(y_{mixup}, logits )$;
      }
  }
  
  \caption{The proposed \ourmodel algorithm.}
  \label{algo:augmentation}
  \end{algorithm}

\begin{figure}[h]
  \centering {
    \includegraphics[width=1.0 \linewidth]{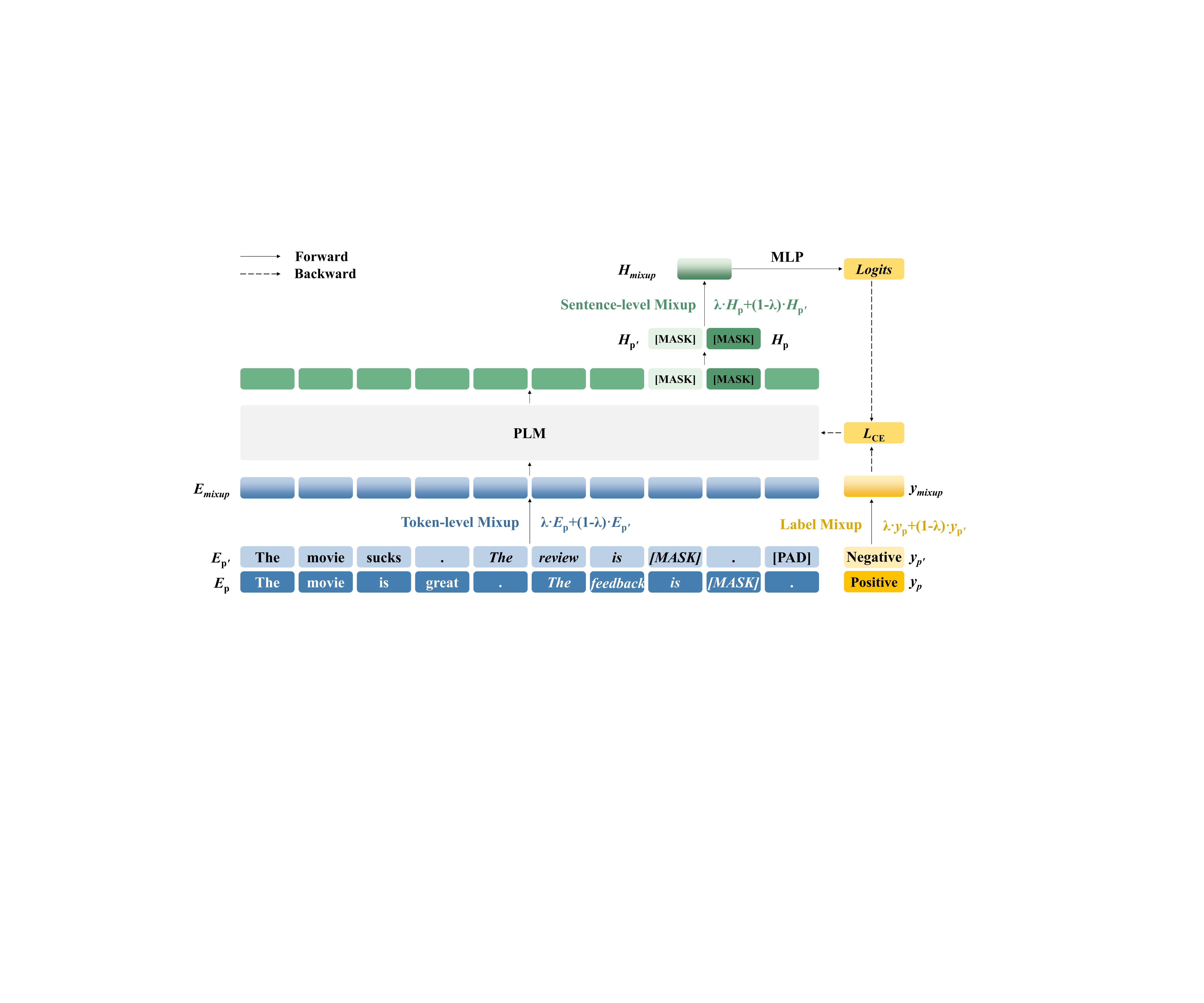}
  }
  \caption{
  The illustration of \ourmodel. 
  The \textbf{token-level Mixup} interpolates word embeddings of the original and augmented prompts as model inputs.
  The \textbf{sentence-level Mixup} interpolates hidden vectors at two [MASK] positions for prediction, while the original and augmented labels are mixed for loss calculation.
  $\lambda$ is a hyperparameter setting the Mixup ratio across token-level, sentence-level, and label Mixup.
  }
  \label{fig:main_method}
\end{figure}

\section{Approach: \ourmodel}\label{Sec.Approach}
\subsection{Overview}

In this section, we introduce \ourmodel, a simple yet effective DA method for prompt-based learning to boost performance and efficiency.
To begin with, we augment the original prompt $\mathbf{p}$ by creating label-preserving or label-flipping prompts $\mathbf{p}'$, as described in Section~\ref{Sec:background-mixup}.\footnote{We collectively refer to $\mathbf{p}_p$ and $\mathbf{p}_f$ in Sec.~\ref{Sec:background-mixup} and Fig.~\ref{fig:DA_example} as $\mathbf{p}'$.}
Moreover, to provide a more comprehensive task representation and better train the model using $\mathbf{p}$ and $\mathbf{p}'$, we propose a three-level Mixup strategy that includes the token-level, the sentence-level, and the template-level Mixup.
We summarize the proposed \ourmodel in Algorithm~\ref{algo:augmentation}, and describe it in detail below.

\subsection{Token-level Mixup}\label{sec:token-level}

The token-level Mixup interpolates the word embeddings of the original prompt and the augmented prompt to obtain new virtual sample representations as model inputs. Fig.~\ref{fig:main_method} illustrates this process.

To start, we obtain the word embeddings for both the original prompt $\mathbf{p}$ and the augmented prompt $\mathbf{p}'$. The word embeddings for each token are constructed by summing its corresponding token, segment, and position embeddings. Specifically, we can express these embeddings as:
\begin{eqnarray}
  E_{\mathbf{p}} =  tok_{\mathbf{p}} + seg_{\mathbf{p}} + pos_{\mathbf{p}},\label{equ:emb_origin_input}
  \\
  E_{\mathbf{p'}} =  tok_{\mathbf{p}'} + seg_{\mathbf{p}'} + pos_{\mathbf{p}'},\label{equ:emb_aug_input}
  \\
  E_{mixup} =  \lambda \cdot E_{\mathbf{p}} + (1-\lambda) \cdot E_{\mathbf{p}'},\label{equ:emb_interpolation}
\end{eqnarray}
where $tok_{\mathbf{p}}$, $seg_{\mathbf{p}}$, and $pos_{\mathbf{p}}$ denote the token, segment, and position embeddings of the original prompt $\mathbf{p}$. $tok_{\mathbf{p}'}$, $seg_{\mathbf{p}'}$, and $pos_{\mathbf{p}'}$ represent those of the augmented prompt $\mathbf{p}'$.
The word embeddings of original, augmented, and interpolated prompts are $E_{\mathbf{p}}$, $E_{\mathbf{p}'}$, and $E_{mixup}$.
The Mixup ratio $\lambda$ is drawn from a Beta distribution $\lambda \sim \beta(\alpha, \alpha)$ (Equation~\ref{equ:mixup}).
As the hyperparameter $\alpha$ approaches zero, $E_{mixup}$ becomes very similar to either $E_{\mathbf{p}}$ or $E_{\mathbf{p}'}$.
Conversely, as $\alpha$ approaches $+ \infty$, $E_{mixup}$ approaches the midpoint between $E_{\mathbf{p}}$ and $E_{\mathbf{p}'}$. We employ the mixed $E_{mixup}$ to train our model.

The token-level Mixup, as illustrated in Fig.~\ref{fig:main_method}, is capable of handling inputs of varying lengths. We also investigated an alternative implementation that aligns two inputs at the [MASK] position, with performance generally comparable to the current approach. 
Given the existence of the sentence-level Mixup in our method, we adopt the current token-level Mixup implementation, which uniformly aligns inputs to the left~\citep{cheng2020advaug[25]}.

\begin{figure}[h]
  \centering {
    \includegraphics[width=1.0 \linewidth]{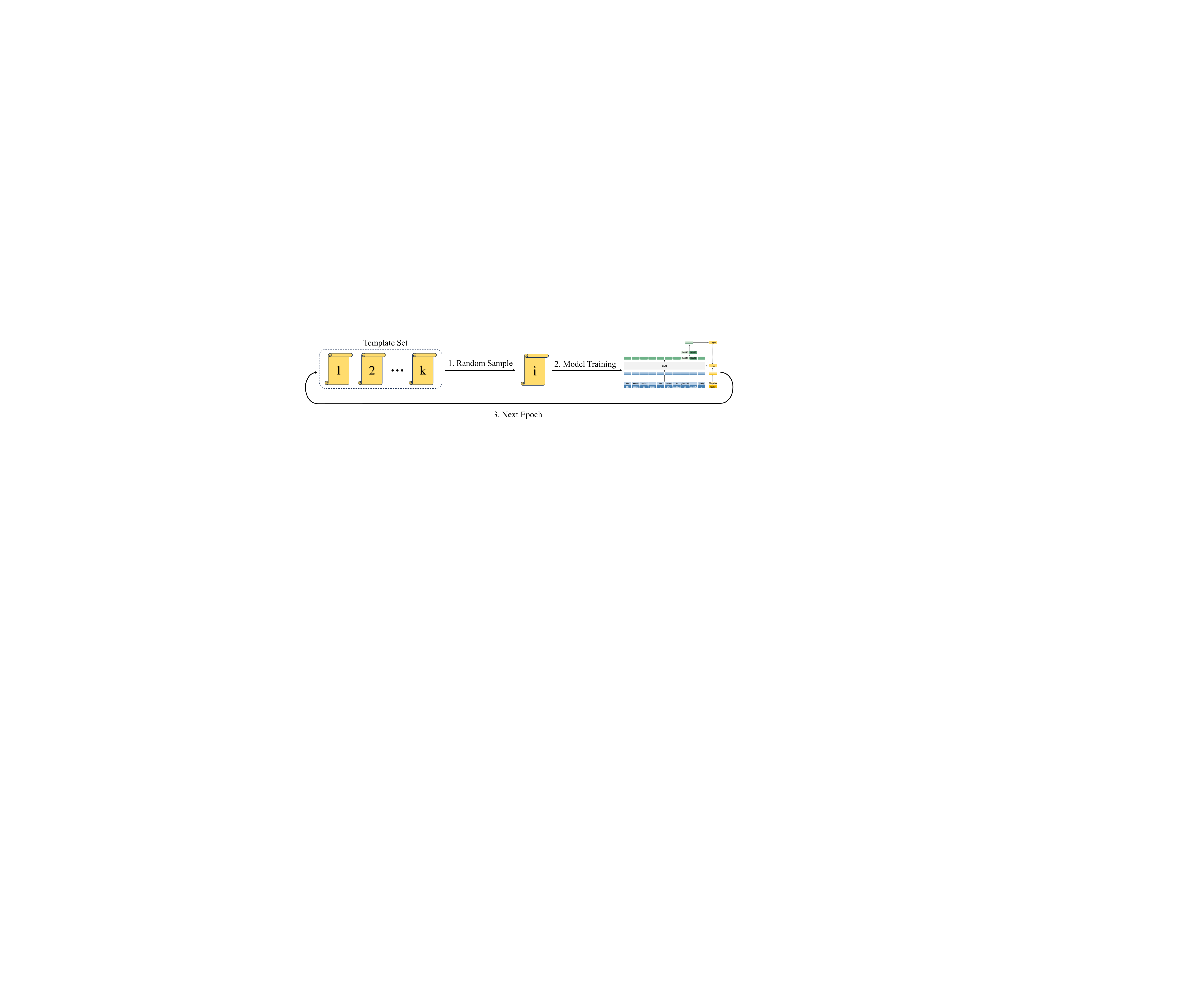}
  }
  \caption{
  In training, the \textbf{template-level Mixup} utilizes various templates to train a single model.
  At the beginning of each epoch, a random template from the template set is chosen for the model training of that epoch.
  }
  \label{fig:epoch_level}
\end{figure}

\subsection{Sentence-level Mixup}

The sentence-level Mixup merges hidden vectors at [MASK] positions from the sequence representations of the model input for prediction (Fig.~\ref{fig:main_method}).
In prompt-based learning, PLMs map these [MASK] representations to task labels, implying these hidden states offer critical insights about the task.

To leverage this information, we interpolate the hidden vectors at the [MASK] positions and the labels 
with Mixup ratio $\lambda$ from a Beta distribution $\lambda \sim \beta(\alpha, \alpha)$. This process can be detailed as:
\begin{eqnarray}
  H_{mixup} =  \lambda \cdot H_{\mathbf{p}} + (1-\lambda) \cdot H_{\mathbf{p}'},\label{equ:mask_mixup} \\
  y_{mixup} = \lambda \cdot y_p + (1 - \lambda) \cdot y_{p'}. \label{equ:label-mixup}
\end{eqnarray}

First, we encode the input representations $E_{mixup}$ from Section~\ref{sec:token-level} and extract hidden vectors $H_{\mathbf{p}}$ and $H_{\mathbf{p}'}$ at the [MASK] positions.
Using $\lambda$, we interpolate these to get $H_{mixup}$ for prediction and also interpolate the original and augmented labels $y_p$ and $y_{p'}$ for loss computation. 
Lastly, an MLP layer computes the logits, determining the cross-entropy loss between $y_{mixup}$ and model logits.
\begin{eqnarray}
  Logits = \mbox{MLP}(H_{mixup}), \\
  \mathcal{L} \sim L_{CE}(y_{mixup}, Logits).
\end{eqnarray}

\subsection{Template-level Mixup}\label{method.template mixing-strategy}

During training, the template-level Mixup uses various templates for prompt creation, enhancing the model's learning from diverse sources.
This strategy also boosts inference efficiency compared to ensembling methods trained with separate templates~\citep{schick-schutze-2021-exploiting}. 
Fig.~\ref{fig:epoch_level} illustrates the template-level Mixup.

The previously mentioned token-level and sentence-level Mixup strategies use a single template during training. Similarly, PET trains multiple models, each with its unique template. During inference,
the ensemble predictions of these models are combined to produce the final result.
However, we believe that training with one template per model might lead models to favor 
\textit{memorization} rather than \textit{generalization}
in few-shot prompt-based learning~\citep{ZhangCDL18_mixup}.

To overcome the constraints of using one template per model, we introduce the template-level Mixup, utilizing various templates for prompt creation in training.
For each epoch, we randomly select a template from the template set $T$ and pair it with all input samples to construct corresponding prompts as model inputs.
This ensures that each input text in the dataset is exposed to all available templates, allowing the model to holistically obtain information from the corresponding prompts. 
During training, the token-level and the sentence-level Mixup techniques are applied to the prompts.

The template-level Mixup also improves the efficiency of model inference.
In the backbone PET approach, each template is used to train a separate model, and during prediction, the corresponding ensemble prediction from multiple models is combined to produce the final result.
In contrast, \ourmodel comprehensively integrates the information from multiple templates by training a single model on all available templates.
This allows for efficient and streamlined inference since only one model needs to be used for prediction.
Specifically, assuming that a dataset has $n$ templates, the time cost of \ourmodel during inference is only $1/n$ of that of the backbone PET approach.
This demonstrates the potential for \ourmodel to improve the efficiency of prompt-based learning without sacrificing performance.

\section{Experiments}\label{Experiments}

We perform experiments on FewGLUE~\citep{schick-schutze-2021-just}, a widely used few-shot benchmark~\citep{zhou2021flipda}.
Our results demonstrate that the proposed \ourmodel is effective in improving few-shot performance by constructing an interpolated Mixup representation between the original prompt and the augmented prompt.

\begin{table}[t]
\centering
\setlength\tabcolsep{4pt} 
\adjustbox{width=0.55\linewidth}{

\begin{tabular}{lccccc}
\toprule
\textbf{Statistics}           & \textbf{CB}   & \textbf{RTE}  & \textbf{BoolQ} & \textbf{WSC}  & \textbf{MultiRC} \\
\midrule
Metrics  & Acc. / F1    & Acc.    & Acc.     & Acc.    &  EM / F1$_\text{a}$       \\
\# Train & $32$    & $32$    & $32$     & $32$    & $32$      \\
\# Dev                     & $57$  & $278$  & $3,270$   & $104$  & $953$     \\
\# Templates                         & $3$    & $4$    & $6$     & $3$    & $3$       \\
\bottomrule
\end{tabular}
}
 \caption{Statistics for the datasets. \textit{\# Train}, \textit{\# Dev}, and \textit{\# Templates} denote sample counts in the training, development sets, and template numbers in the dataset, respectively.}
\label{tab:datasets}

\end{table}

\subsection{Experiment Setup}
\subsubsection{Tasks}\label{Exp.tasks}

FewGLUE is the few-shot version of SuperGLUE~\citep{DBLP:conf/nips/WangPNSMHLB19}, including a diverse set of natural language understanding tasks, such as question answering (BoolQ and MultiRC)\citep{DBLP:conf/naacl/ClarkLCK0T19,DBLP:conf/naacl/KhashabiCRUR18}, word sense disambiguation (WSC)\citep{DBLP:conf/kr/LevesqueDM12}, and textual entailment (CB and RTE) \citep{Marneffe2019TheCI,DBLP:conf/mlcw/DaganGM05,BarHaim2006TheSP}. Each task includes a 32-sample train set and a validation set. Successfully solving these tasks requires a deep understanding of natural language.\footnote{There are three additional datasets in FewGLUE. The experimental results of WIC~\citep{DBLP:conf/naacl/PilehvarC19} is close to random. The task forms of ReCoRD~\citep{Zhang2018ReCoRDBT} and COPA~\citep{DBLP:conf/aaaiss/RoemmeleBG11} are not well suited for prompt-based augmentation. Thus, we do not conduct experiments on them.} For more detailed information, please refer to Table~\ref{tab:datasets}.

\subsubsection{Baselines}\label{sec:baseline}
We integrate a comprehensive selection of nine augmentation methods, serving as baselines for augmenting the PET backbone. 
We will introduce these methods in the following, and then compare them with the proposed \ourmodel from five perspectives in Table~\ref{tab:model-compare}.

\begin{itemize}

  \item \textbf{Synonym Replacement} (Synonym, \citep{DBLP:conf/nips/ZhangZL15}) uses a synonym dictionary called WordNet~\citep{miller1995wordnet} to randomly replace the original words with their synonyms.

  \item \textbf{GloVe Replacement} (GloVe,~\citep{DBLP:conf/emnlp/WangY15}) substitutes the original words with nearby words from pre-trained GloVe embeddings~\citep{pennington-etal-2014-glove}.

  \item \textbf{Easy Data Augmentation} (EDA) performs operations like synonym replacement, random insertion, random swap, and random deletion.\footnote{The implementation of \textit{synonym replacement} is based on \cite{DBLP:conf/nips/ZhangZL15} above.}

  \item \textbf{Back Translation} (BT,~\citep{xie2019unsupervised[9]}) translates the original text into other languages and then back. The resulting text is combined with the original.\footnote{
  We get the augmented data with 9 intermediate languages in BT-10, and with 5 in BT-6.
  }

  \item \textbf{TinyBERT}~\citep{jiao-etal-2020-tinybert} randomly replaces the original token in the input text with either word predicted by a \textit{Bert-base-cased} model~\citep{devlin2018bert} (for singlepiece word) or words derived by GloVe~\citep{pennington2014glove} (for multiple-piece word).

  \item \textbf{T5-MLM}~\citep{DBLP:journals/jmlr/RaffelSRLNMZLL20} randomly masks some tokens in the original input text and fills the blanks with T5 model in a template-based data cloze.

  \item \textbf{Mixup}~\citep{guo2019augmenting[24],cheng2020advaug[25]} adopts the vanilla strategy in Equ.~\ref{equ:mixup}-\ref{equ:mixup-label} and interpolates one original input text with another random input text in the training set.

  \item \textbf{FlipDA}~\citep{zhou2021flipda} constructs both label-preserving and label-flipping augmented data given the original input text. The implementation is based on the T5-MLM above. It additionally applies data filtering to the augmented data.

\end{itemize}

\begin{table}[t]
\centering
\setlength\tabcolsep{6pt} 
\adjustbox{width=0.85\linewidth}{

\begin{tabular}{lccccc}
\toprule
\textbf{Methods}           & \textbf{Object} & \textbf{Ext.Res}   & \textbf{Pres / Flip}  & \textbf{Mode} & \textbf{Level}  \\
\midrule
  Synonym  & Input & \checkmark & Pres  & Text  & Token \\
  GloVe & Input & \checkmark & Pres & Text  & Token  \\
  EDA & Input  & \checkmark & Pres & Text  & Token   \\
  BT-10 & Input & \checkmark & Pres  & Text  & Sent \\
  BT-6  & Input & \checkmark & Pres  & Text  & Sent  \\
  TinyBERT    & Input & \checkmark & Pres  & Text  & Token \\
  T5-MLM & Input   & \checkmark & Pres  & Text  & Token \\
  Mixup & Input  & - & Pres \& Flip  & Emb  & Sent \\
  FlipDA    & Input & \checkmark & Pres \& Flip  & Text & Token \\
\midrule
  \textbf{\ourmodel}  & Input \& Tmpl & \checkmark & Pres \& Flip & Text \& Emb & Token \& Sent \& Tmpl \\
\bottomrule
\end{tabular}
  }
 \caption{
  Comparison of \ourmodel with baseline methods. 
  \textit{Object} indicates the target of augmentation. 
  \textit{Tmpl} represents the template. 
  \textit{Ext.Res} signifies if methods utilize external resources beyond the dataset, such as synonyms, embeddings, or additional models. 
  \textit{Pres / Flip} highlights whether the methods use label-preserving (\textit{Pres}) or label-flipping (\textit{Flip}) augmentation. 
 \textit{Mode} indicates whether the augmentation is applied on the text itself or its underlying embeddings (\textit{Emb}).
  \textit{Level} shows how broadly the augmentation is applied, whether it is on individual tokens, entire sentences (\textit{Sent}), or templates (\textit{Tmpl}).
  }
  \label{tab:model-compare}

\end{table}

In this paper, we propose \ourmodel, a simple yet effective DA method for prompt-based learning.
\ourmodel draws additional virtual examples through three-level Mixup based on the label-preserving and label-flipping augmented prompts to expand the support of the training distribution.
To evaluate the effectiveness and capabilities of \ourmodel, we undertake a comprehensive comparison against the above baselines from five perspectives. Detailed results of this comparison can be found in Table~\ref{tab:model-compare}.

\ourmodel optimizes the above baselines from various criteria:
\tif{(1) Object:}
\ourmodel augments the templates, enriching the model's understanding of the task while simultaneously reducing its heightened sensitivity to specific templates.
\tif{(2) Ext.Res:}
\ourmodel employs T5 as an external resource, which generates high-quality augmented prompts for the subsequent three-level Mixup.
\tif{(3) Pres / Flip:} 
Label-flipping augmented data provides useful information for classification and largely improves generalization~\citep{zhou2021flipda}.
\tif{(4) Mode:}
\ourmodel adopt the Mixup strategy~\citep{ZhangCDL18_mixup,ZhangCDL18Mixup} to draw additional \textit{virtual} examples from the vicinity of the augmented prompt for a comprehensive distribution~\citep{ZhangCDL18Mixup}. 
\tif{(5) Level:}
\ourmodel adopts a three-level Mixup to achieve multi-dimensional augmentation, aiming for enhanced performance and inference efficiency.

Overall, our proposed \ourmodel is more \textbf{comprehensive} than the baselines and presents significant advancements in few-shot prompt-based learning.
By augmenting the entire prompt, it successfully reduces the model's sensitivity to template variations.
By a three-level Mixup with label-flipping and label-preserving prompts, it enhances performance and inference efficiency.
Details are in Section~\ref{sec:main-results}.

\begin{table}[t]
  \centering
  \setlength\tabcolsep{2.5pt} 
  \adjustbox{width=0.55\linewidth}{
  
  \begin{tabular}{lccccc}
  \toprule
  \textbf{Hyperparameters}           & \textbf{CB}   & \textbf{RTE}  & \textbf{BoolQ} & \textbf{WSC}  & \textbf{MultiRC} \\
  \midrule
  \texttt{batch\_size}  & $2$    & $2$    & $2$     & $4$    & $1$       \\
  \texttt{grad\_acc\_steps} & $8$    & $8$    & $8$     & $4$    & $16$      \\
  \texttt{max\_seq\_length}              & $256$  & $256$  & $256$   & $128$  & $512$     \\
  \texttt{max\_steps}                    & $250$  & $250$  & $250$   & $250$  & $250$     \\
  \texttt{adam\_epsilon}                      & $1e-8$ & $1e-8$ & $1e-8$  & $1e-8$ & $1e-8$    \\
  \texttt{learning\_rate}                     & $1e-5$ & $1e-5$ & $1e-5$  & $1e-5$ & $1e-5$    \\
  \texttt{max\_grad\_norm}                    & $1.0$  & $1.0$  & $1.0$   & $1.0$  & $1.0$     \\
  \texttt{weight\_decay}                      & $0.01$ & $0.01$ & $0.01$  & $0.01$ & $0.01$    \\
  \texttt{mixup\_alpha}                       & $0.5$  & $0.5$  & $0.1$   & $0.1$  & $0.5$     \\
  \bottomrule
  \end{tabular}
  }
    \caption{Hyperparameters used in \ourmodel.}
    \label{tab:hyperparameters}
  
  \end{table}

\subsubsection{Implementation Details}\label{sec:experiment-parameters}

From Section~\ref{Sec:background-mixup}, we augment prompts and include both \textit{label-preserving} and \textit{label-flipping} types. 
The input text and templates are augmented separately. We generate label-preserving and label-flipping augmented text, but label-preserving templates only.
Note that we do not generate label-flipping templates to ensure the correctness and controllability of the label corresponding to the augmented prompt. 
Following~\cite{zhou2021flipda}, we use a cloze pattern to combine the input text (or template) and label into a single sequence, and mask a fixed percentage of the tokens. A pre-trained T5 model~\citep{Raffel2019ExploringTL} is used to fill in the blanks and generate an augmented sample.

\begin{table}[t]
  \centering
  \adjustbox{width=0.8\linewidth}{

  \begin{tabular}{lcccccccccc}
  \bottomrule

   & \multicolumn{2}{c}{\textbf{CB}} & \textbf{RTE} & \textbf{BoolQ}  & \textbf{WSC} & \multicolumn{2}{c}{\textbf{MultiRC}}  & \textbf{Avg.} \\
  Method & Acc. & F1 & Acc. & Acc. & Acc. & EM & F1$_\text{a}$  & -    \\
  \midrule
  \textbf{PET~~~~~~~~~~~~} & 82.74  & 74.84  & 61.40 & 72.47 & 77.03    & 33.04  & 74.64  & 68.02 \\
  ~~+ Synonym  & \underline{83.33}  & \underline{78.12}  & 59.24  & \underline{74.98}    & \underline{78.74}    & \underline{34.09}  & \underline{75.55} & \underline{69.15}\\
  ~~+ GloVe & 82.14  & 74.39  & \underline{61.91}  & \underline{74.51}  & 75.00       & 32.72  & \underline{75.20}  & 67.98  \\
  ~~+ EDA & 81.10 & 73.58  & 58.33  & \underline{72.86}    & 75.85    & 28.74  & 73.05 &  66.22\\
  ~~+ BT-10 & 82.44  & \underline{77.72}  & 55.93  & \underline{74.59}     & -        & 32.06  & \underline{74.69} & 66.24 \\
  ~~+ BT-6  & \underline{82.89}  & \underline{76.55}  & 57.46  & \underline{75.36}   & -        & \underline{34.85}  & \underline{75.82} & 67.16 \\
  ~~+ TinyBERT    & \underline{85.42}  & \underline{82.35}  & 58.66  & \underline{72.60}  & \underline{78.95}    & 30.47  & 73.20 & \underline{68.81} \\
  ~~+ T5-MLM & \underline{83.48}  & \underline{75.01}  & \underline{62.27}  & \underline{73.86}  & \textbf{79.17}  & \underline{33.79}  & 74.06 & \underline{68.81} \\
  ~~+ Mixup & \underline{83.93}  & \underline{79.28}  & \underline{62.06}  & \underline{75.03}   & 68.70     & \underline{34.06}  & \underline{74.66}  & \underline{68.25} \\
  ~~+ FlipDA    & \underline{86.31}  & \underline{82.45}  & \underline{70.67} & \underline{76.98} & \underline{78.74}    & \underline{36.38}  & \underline{76.23} & \underline{72.54}\\
\midrule
  ~~+ \textbf{\ourmodel}  & \textbf{86.94}  & \textbf{82.67}  & \textbf{71.16}  & \textbf{78.96}  &  \underline{79.13} &  \textbf{36.41} & \textbf{76.44} & \textbf{73.10}\\
\bottomrule

  \end{tabular}
}
  \caption{Performance of baseline methods and \ourmodel based on the backbone PET. \underline{Underline} denotes values that outperform PET. \textbf{Bold} denotes the best-performed ones of the task. “Avg.” is the average of scores.}
  \label{tab:main-result}

\end{table}

For model training, we adopt the experimental settings used in PET~\citep{schick-schutze-2021-just} and conduct a grid search (see Table~\ref{tab:hyperparameters}).
We set the batch size to 2 for CB, RTE, and BoolQ, to 4 for WSC, and to 1 for MultiRC.
We configure the gradient accumulation steps to 8 for CB, RTE, and BoolQ, to 4 for WSC, and to 16 for MultiRC.
We set the max sequence length for tokenized inputs to 256 for CB, RTE, and BoolQ, to 128 for WSC, and to 512 for MultiRC.
Across all tasks, we consistently set the max steps to 250, adam epsilon to \(1 \times 10^{-8}\), and learning rate to \(1 \times 10^{-5}\).
To prevent gradients from becoming too large, we set the max grad norm to 1.0 for all tasks, and for regularization, we set the weight decay to \(0.01\) across all tasks.
Lastly, pivotal for the three-level Mixup, we set the mixup alpha ($\lambda$ in Equ.~\ref{equ:emb_interpolation}-\ref{equ:label-mixup}) to \(0.5\) for CB, RTE, and MultiRC, and to \(0.1\) for BoolQ and WSC.

To evaluate the effectiveness of DA methods, we augment the backbone PET~\citep{schick-schutze-2021-just} with \ourmodel and other DA baselines.
We utilize \textit{Albert-xxlarge-v2}\citep{DBLP:conf/iclr/LanCGGSS20} as the PLM and measure the performance using the identical metrics in Table~\ref{tab:datasets}, namely Acc., F1, F1$_\text{a}$, and EM.
Since few-shot learning typically exhibits significant performance fluctuations~\citep{DBLP:journals/corr/abs-2002-06305,schick-schutze-2021-just}, we use five independent seeds and report their average performance. All of our experiments were performed on a Linux platform equipped with NVIDIA A100 (40G).

\subsection{Main Results}\label{sec:main-results}

In this paper, we conduct experiments on five different datasets, namely CB, RTE, BoolQ, WSC, and MultiRC.
We compare \ourmodel with the backbone PET and nine augmentation baselines from Sec.~\ref{sec:baseline}, analyzing the results in three dimensions.
Besides performance improvements, it's essential to consider the computational cost associated with these enhancements. Thus, we further analyze the computational implications of these improvements.

Table~\ref{tab:main-result} presents a detailed overview of our experimental findings.\footnote{The results of baselines are obtained from~\cite{zhou2021flipda} using the same experimental settings as ours.}
\tif{(1)} Overall performance: \ourmodel achieved the highest average performance of $73.10$. It not only outshines the backbone PET by $5.08\%$, but also surpasses the best baseline FlipDA by $0.56\%$;
\tif{(2)} Performance improvement beyond the backbone:
when compared against the backbone PET, \ourmodel consistently outperforms it across all datasets, whereas other baseline methods, except for FlipDA, were only effective on a few tasks (denoted with underlines);
\tif{(3)} Performance comparison with baselines across datasets:
In terms of inter-dataset performance, we compare the performance of all models in each dataset. \ourmodel is the best performing model on all datasets (except for WSC, where \ourmodel is the second-best method), and consistently outperforms the strongest baseline, FlipDA, on all datasets.

Based on the experimental results presented above, we argue that there are several \textbf{challenges} in few-shot prompt-based learning augmentation.
Specifically, models are highly sensitive to the choice of templates, where small changes can lead to notable declines in effectiveness~\citep{gao2020making,vandeKar2022DontPS,cao-etal-2022-prompt}.
Additionally, existing methods often employ model ensembles across different templates and limit the model's inference efficiency.~\citep{schick-schutze-2021-just}.
Consequently, some traditional augmentation methods still face challenges in model performance and inference efficiency within few-shot prompt-based learning.

In comparison to existing methods, our proposed \ourmodel leverages three-level Mixup strategies to sample virtual examples from different perspectives. 
By combining information from various templates, we further enhance the comprehensiveness of the model's understanding.
This intricate design ensures model effectiveness and robustness across different tasks.
The excellent results of \ourmodel in our experiments demonstrate its ability to address the aforementioned challenges stably.

Meanwhile, our DA method brings a certain increase in computational cost, which we consider is acceptable.
For the average training cost, there is an increase of 20.9\% in training time during each iteration (43s $\rightarrow$ 52s) and an increase of 15.24\% in the number of model parameters (223M $\rightarrow$ 257M) when keeping the batch size constant.
For the average inference cost, \ourmodel improves the efficiency of model inference.
\ourmodel applies multiple templates to train a single model rather than multiple ensemble models as PET does.
Specifically, assuming that a dataset has $n$ templates, the time cost of \ourmodel during inference is only $1/n$ of that of the backbone PET approach.

In summary, the proposed \ourmodel achieves excellent results across five datasets via three-level Mixup. 
As an augmentation method, it experiences a certain increase in computational cost and the number of parameters compared to the backbone.
However, given the overall $5.08\%$ improvement in performance, the decrease in inference time, and robustness across different tasks, we consider these computational cost increases to be within an acceptable range.\footnote{A more detailed exploration of the robustness of \ourmodel can be found in Sec.~\ref{sec:standard-deviation}.}

\begin{table}[t]
  \footnotesize
    \centering
    \adjustbox{width=0.8\linewidth}{

    \begin{tabular}{lcccccccccc}
    \hline
     & \multicolumn{2}{c}{\textbf{CB}} & \textbf{RTE} & \textbf{BoolQ} & \textbf{WSC} & \multicolumn{2}{c}{\textbf{MultiRC}}  & \textbf{Avg.} \\
     Method & Acc. & F1 & Acc. & Acc. & Acc. &  EM & F1$_\text{a}$  & -    \\
    \hline
    \textbf{\ourmodel}  & \textbf{86.94}  & \textbf{82.67}  & \textbf{71.16}  & \textbf{78.96}  &  \textbf{79.13} &  \textbf{36.41} & \textbf{76.44} & \textbf{73.10}\\

    ~~\footnotesize{w/o Token}    & \underline{86.07} & \underline{82.57}   & 60.14 & 78.61 &   \underline{78.46}   & \underline{35.76}  & \underline{76.39} & 71.14 \\
    
    ~~\footnotesize{w/o Sent}   & 85.35 &  80.59  & 67.22 & \underline{78.84} &   77.88    & 35.73 & 76.04  & 71.66 \\
    
    ~~\footnotesize{w/o Tmpl}     & 85.00 &  80.49  & \underline{69.31} &   78.52 & 78.26&   35.39 & 74.76 & \underline{71.68} \\

    \hline
    \end{tabular}
    }
    \caption{ Experiment results of removing the token (\textit{Token}), sentence (\textit{Sent}), or template-level (\textit{Tmpl}) Mixup strategies from \ourmodel. \underline{Underline} denotes values that outperform PET. \textbf{Bold} denotes the best-performed ones of the task. “Avg.” is the average of scores. All results are the average over 5 seeds.}
    \label{tab:ablation-Embmix-Maskmix}
  
  \end{table}

\subsection{Analysis}

In this section, we address the following research questions for a deeper understanding of \ourmodel:

\begin{itemize}
\item [\tif{(1)}] Do the token-level, the sentence-level, or the template-level Mixup strategies make equal contributions to the performance of \ourmodel?

\item [\tif{(2)}] Is it essential to augment both input text and templates concurrently?
\item [\tif{(3)}] Does \ourmodel demonstrate significant performance fluctuations across multiple seeds compared to the backbone PET?
\end{itemize}

\subsubsection{Three-level Mixup: Separate Contributions} \label{sec:mixup-ablation}

To delve deeper into the impact of removing the token-level, the sentence-level, or the template-level Mixup strategies on the overall performance of our model, we systematically conduct ablation experiments and analyze the results.

The experimental results in Table~\ref{tab:ablation-Embmix-Maskmix} demonstrate that:
\tif{(1)} \ourmodel outperforms all other methods on all datasets, indicating that the Mixup strategies at the token, sentence, and template levels all contribute to performance improvement;
\tif{(2)} The impact of \textit{w/o Token} is minimal overall.
While it had a significant negative impact on the RTE dataset (over $10\%$ lower than \ourmodel), leading to a corresponding decrease in average performance, it has minimal impact on most datasets.
This indicates that the token-level Mixup, which samples new distributions at the word embedding level, is relatively \textit{shallow} and, while it can improve model performance, is not as effective as the other two Mixup strategies;
\tif{(3)} The impact of \textit{w/o Sent} and \textit{w/o Tmpl} are similar, with the former having a slightly larger impact on average ($0.02\%$ lower).
These findings suggest that both strategies are effective: the sentence-level Mixup provides a deeper understanding of samples than the token-level Mixup, while the template-level Mixup constructs a comprehensive task distribution by utilizing multiple templates during training.

\begin{table}[t]
\footnotesize
\centering
\adjustbox{width=0.8\linewidth}{
\begin{tabular}{lcccccccccc}
\hline
& \multicolumn{2}{c}{\textbf{CB}} & \textbf{RTE} & \textbf{BoolQ} & \textbf{WSC} & \multicolumn{2}{c}{\textbf{MultiRC}}  & \textbf{Avg.} \\
Method & Acc. & F1 & Acc. & Acc. & Acc. &  EM & F1$_\text{a}$  & -    \\
\hline
\textbf{\ourmodel}  & \textbf{86.94}  & \textbf{82.67}  & \textbf{71.16}  & \textbf{78.96}  &  \textbf{79.13} &  \textbf{36.41} & \textbf{76.44} & \textbf{73.10}\\

~~\footnotesize{w/o Template Aug}    & \underline{84.28}  & \underline{80.59}   & \underline{65.84} & 78.50 &   77.69   & \underline{34.41}  & \underline{75.71} & \underline{71.00} \\

~~\footnotesize{w/o Text Aug}   & 83.71 &  76.70  & 65.56& \underline{78.54} &   \underline{78.08}    & 34.17 &75.18  & 70.28 \\
\hline
\end{tabular}
}
\caption{ Experiment results of removing the augmentation (\textit{Aug}) of vanilla text input or templates from \ourmodel across five datasets. \underline{Underline} denotes values that outperform PET. \textbf{Bold} denotes the best-performed ones of the task. “Avg.” is the average of scores. All results are the average over 5 seeds.}
\label{tab:mixup-range}

\end{table}

\subsubsection{Necessity of Augmenting Text and Templates}

To understand the influence of removing the augmentation for either vanilla input text or the associated templates, we conduct a series of ablation experiments and analyze the results.

\begin{figure}[tbp]

  \centering {
  \adjustbox{width=0.7\linewidth}{
    \includegraphics[width=1.0 \linewidth]{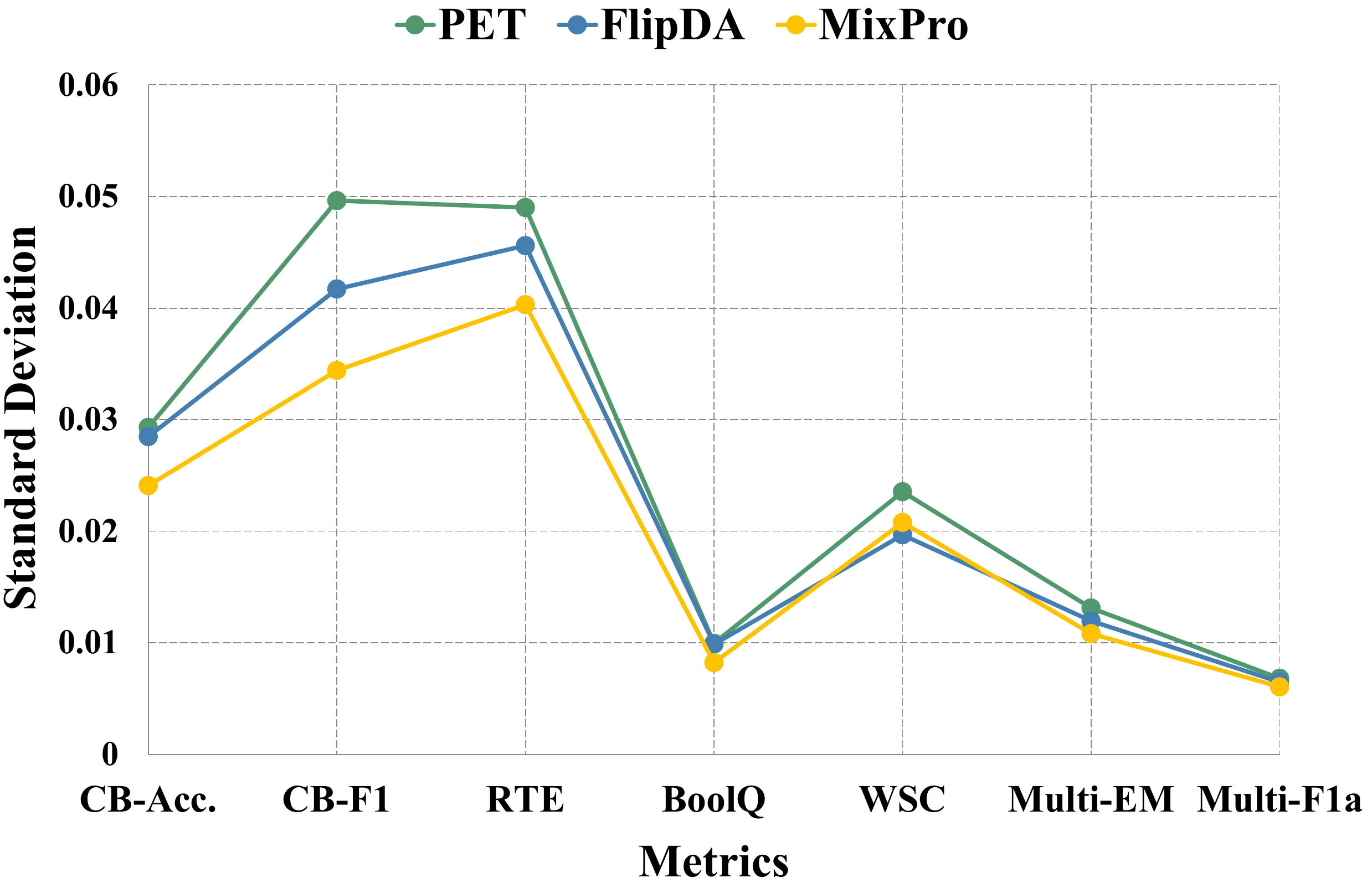}

  }
  }
  \caption{Standard deviation of performance of PET, FlipDA and \ourmodel on five datasets. ``MULTI-EM'' and ``MULTI-F1A''  denote the standard deviations of the EM score and F1$_\text{a}$ score of the models on MultiRc.}
  \label{fig:line}
\end{figure}

The results from the experiments, which are detailed in Table~\ref{tab:mixup-range}, shed light on several key findings:
\tif{(1)} Our proposed method, \ourmodel, consistently outperforms the other two methods across all datasets.
This distinction underscores the significance of augmenting both the vanilla input text and templates. 
\tif{(2)} When examining the effects of \textit{w/o Template Aug} and \textit{w/o Text Aug}, the findings suggest a similar impact across the five datasets.
Moreover, a closer look shows that \textit{w/o Text Aug} has a slightly bigger impact, decreasing performance by an average of $0.72\%$.
This is likely because input texts are usually longer than templates. 
Therefore, augmenting the text provides richer information and achieves better performance than augmenting the templates.

\subsubsection{Measuring the fluctuation of model performance}\label{sec:standard-deviation}

We determine the standard deviation of the results for \ourmodel, FlipDA, and PET using five seeds spread across five distinct datasets.

The experimental results, as illustrated in Fig.~\ref{fig:line}, bring noteworthy observations:
Both augmentation methods, FlipDA and \ourmodel, demonstrate a standard deviation across the five datasets that is markedly lower than the backbone PET.
Moreover, FlipDA exhibits a consistently smaller standard deviation than \ourmodel, with the sole exception being the WSC dataset.

These outcomes underscore the effectiveness of data augmentation (DA) techniques in enhancing a model's robustness and generalizability.
Specifically, \ourmodel, empowered by its three-level Mixup strategy, establishes a more diversified data distribution, leading to superior performance metrics.
By augmenting templates and training the model concurrently on them (see Sec.~\ref{Sec.Approach} and Sec.~\ref{sec:baseline}), \ourmodel minimizes its sensitivity to particular templates. As a result, it achieves a more stable performance, surpassing both the backbone and the strongest baseline across a range of datasets.

\section{Related Works}\label{Sec.RelatedWorks}

\subsection{Prompt-based Learning}

Prompt-based learning converts tasks into cloze questions with templates~\citep{DBLP:journals/corr/abs-2107-13586}. It includes two types: hard and soft prompts, which we will detail separately.

In \textbf{hard prompts}, templates are human-readable text strings, as used in this paper. Two types exist:
\tif{(1)} Manually created prompts based on human introspection, like those in \cite{petroni-etal-2019-language} and \cite{brown2020language} to probe LMs, or predefined ones in few-shot learning~\citep{schick2020fewshot, schick2021exploiting, schick2021its}.
\tif{(2)} Automatically searched templates from natural language phrases, including methods like prompt mining~\citep{jiang-etal-2020-know}, prompt paraphrasing~\citep{yuan2021bartscore, haviv-etal-2021-bertese}, gradient-based search~\citep{DBLP:conf/emnlp/WallaceFKGS19,autoprompt:emnlp20}, and prompt scoring~\citep{DBLP:conf/emnlp/DavisonFR19}.

\textbf{Soft prompts} use the embedding space of language models, not limited to human-interpretable language.
The advent of continuous prompts has removed the restriction that templates must be parameterized solely by pre-trained language model parameters. 
Instead, templates can have their own parameters that are adjustable based on training data from the downstream task.
Common methods include prefix tuning \citep{li2021prefix,lester2021power,DBLP:journals/corr/abs-2106-13884}, tuning initialized with discrete prompts \citep{zhong2021optiprompt,qin-eisner-2021-learning,Hambardzumyan2021WARPWA}, and hard-soft prompt hybrid tuning \citep{liu2021ptuning,han2021ptr}.

This paper focuses on \textbf{hard prompts} due to their simplicity~\citep{brown2020language}, user-friendliness~\citep{schick2020fewshot}, and competitive performance~\citep{Wen2023HardPM}.
Building on~\cite{schick-schutze-2021-just}, \ourmodel integrates automatic augmentation to enhance performance, robustness, and inference efficiency.
It is lightweight and requires minimal manual input.\footnote{
Manual annotation \citep{petroni-etal-2019-language,brown2020language} is expert-intensive and time-costly,
while automatic generation needs ample data, unsuited for few-shot tasks~\citep{zhou2021flipda}. We will apply data augmentation methods to them for further improvement in the future.
}

\subsection{Data Augmentation}

Data augmentation aims at producing synthetic training data in scenarios with limited data. As prompt-based learning evolves, there are some studies investigating their integration with data augmentation. In this section, we offer a concise overview of these studies.

Data augmentation in prompt-based learning falls into two categories: \tif{(1)} \textbf{Augmenting prompts}:
FlipDA~\citep{zhou2021flipda} generates data using word substitution based on a pre-trained language model and uses a classifier to select label-flipped data.
PromptDA~\citep{Chen2022PromptDALD} derives multiple label words for each class to enrich the label semantic space, while RAPT~\citep{Chowdhury2022NoveltyCP} augments soft prompts.
\tif{(2)} \textbf{Using prompts for augmentation}: 
PromDA~\citep{Wang2022PromDAPD} 
uses soft prompts to generate augmented data for low-resource NLU tasks.
WeakDap~\citep{Chen2022WeaklySD} explores few-shot data augmentation for dialogue understanding by prompting pre-trained language models,
and AUG-FedPrompt~\citep{Cai2022AUGFedPromptPF} annotates unlabeled data via a prompt-based federated learning algorithm.

\textit{Our paper focuses on the first type, \textbf{augmenting prompts}},
and our \ourmodel distinguishes itself from prior work in the following ways: 
\tif{(1)} It augments the complete prompt to enhance performance, and \tif{(2)} it specifically augments hard prompts.
We detail these distinctions below.
\tif{(1)} Most existing methods augment segments of a prompt rather than its entirety.
For instance, FlipDA~\citep{zhou2021flipda} augments only the vanilla input text, bypassing the templates.
Thus, it does not truly augment in the context of prompt-based learning.
However, templates play a vital role in prompt-based learning, and their augmentation proves crucial as shown in Tab.\ref{tab:mixup-range}.
Similarly, PromptDA\citep{Chen2022PromptDALD} focuses only on label augmentation, limiting its diversity.
\tif{(2)} Some research centers on augmenting soft prompts, diverging from our hard prompt augmentation~\citep{Chowdhury2022NoveltyCP}. 
Conversely, \ourmodel augments both the vanilla input text and hard templates using the three-level Mixup. It facilitates the sampling of new virtual distributions during training, leading to excellent performance.

\section{Conclusion and Future Work}\label{Sec.Conclusion}

In this paper, we present a data augmentation method called \ourmodel that is specifically designed for prompt-based learning.
Our method employs a three-level Mixup strategy to generate augmented prompts and comprehensively constructs virtual examples from the vicinity distribution of the original prompts.
Our experiments on five different few-shot learning datasets demonstrate that \ourmodel leads to a substantial improvement in the backbone PET by an average of $5.08\%$.
Furthermore, our method achieves better results with improved efficiency during inference compared to the backbone.

In the future, we plan to further explore two directions.
First, we will improve the model architecture by simultaneously feeding the original and augmented prompts into the model and performing Mixup at the self-attention layer in the PLMs to achieve a more natural and deeper information interaction.
Second, we will conduct more experiments to apply data augmentation to both manually annotated and automatically generated prompts, with the goal of enhancing the performance of both.

\section*{Limitations}

While deploying our method to various downstream tasks, certain limitations emerge concerning the hyperparameter tuning process.
The selection of $\alpha$ plays a pivotal role and visibly influences model performance.\footnote{$\lambda$ denotes the Mixup ratio, sampled from a Beta distribution $\lambda \sim \beta(\alpha, \alpha)$, where $\alpha$ acts as a crucial hyperparameter governing the extent of Mixup.}
\textbf{On the brighter side, we find the process of fine-tuning $\alpha$ to be relatively straightforward and feasible in terms of computational resources.
}
In alignment with prior research~\citep{guo2019augmenting[24],cheng2020advaug[25]}, we select $\alpha$ from a pre-defined, restricted set ($1e-2$, $1e-1$, $5e-1$, and $1$).
Empirical evidence from our experiments across five different datasets distinctly shows a preference for $1e-1$ and $5e-1$, underlining a manageable scope of tuning that brings tangible benefits to model performance.
Furthermore, the entire experimental setup boasts efficiency, seamlessly running on a single V100 GPU (32G), thereby ensuring that the hardware demands remain minimal and accessible. In summation, while the noted limitation exists, it is well within the bounds of manageability, and addressing it opens avenues for enhancing model performance.

Concurrently, our approach inherits certain constraints from FlipDA~\citep{zhou2021flipda}. As depicted in Sec.~\ref{sec:main-results}, while \ourmodel demonstrates remarkable improvements over existing baseline augmentation methods in most scenarios, its performance on the WSC task slightly trails behind some baselines. This is attributed to the complexity of the WSC task, which necessitates the disambiguation of multi-token word senses, a challenge for T5 when it comes to generating label-flipping augmented prompts. The T5 model struggles with fabricating similar entities absent from the original sentence, hindering its ability to generate the required candidate examples. Addressing this issue, particularly devising a more adept pattern-based cloze algorithm for such tasks, remains an area for future exploration~\citep{Rosset2020KnowledgeAwareLM}.

\clearpage
\bibliographystyle{icml2020_url}
\vspace{0pt}
\bibliography{reference}

\end{document}